\documentclass[journal,onecolumn]{IEEEtran}
\ifCLASSINFOpdf
\else
\fi
\usepackage{apacite}
\newcommand{\bx}{\mathbf{x}}
\newcommand{\bv}{\mathbf{v}}

\usepackage{amsmath,amssymb,amsfonts}
\usepackage{algorithmic}
\usepackage{algorithm}

\usepackage{caption}
\usepackage{graphicx}
\usepackage{setspace}
\usepackage{tikz,soul}
\usepackage{lineno}

\begin{document}
%
\title{Adapting the Exploration-Exploitation Balance in Heterogeneous Swarms: Tracking Evasive Targets}

%
%
%

\author{Hian~Lee~Kwa, 
        Victor~Babineau,
        Julien~Philippot,
        Roland~Bouffanais
\thanks{H. L. Kwa is with the Singapore University of Technology and Design and Thales Solutions Asia}
\thanks{V. Babineau, J. Philippot \& R. Bouffanais are with the University of Ottawa}
\thanks{H. L. Kwa is funded by Thales Solutions Asia Pte Ltd, under the Singapore Economic Development Board Industrial Postgraduate Programme (IPP)}
\thanks{Corresponding Author: \texttt{hianlee\_kwa@mymail.sutd.edu.sg}}}

\maketitle

\begin{abstract}
There has been growing interest in the use of multi-robot systems in various tasks and scenarios. The main attractiveness of such systems is their flexibility, robustness, and scalability. An often overlooked yet promising feature is system modularity, which offers the possibility to harness agent specialization, while also enabling system-level upgrades. However, altering the agents' capacities can change the exploration-exploitation balance required to maximize the system's performance. Here, we study the effect of a swarm's heterogeneity on its exploration-exploitation balance while tracking multiple fast-moving evasive targets under the Cooperative Multi-Robot Observation of Multiple Moving Targets framework. To this end, we use a decentralized search and tracking strategy with adjustable levels of exploration and exploitation. By indirectly tuning the balance, we first confirm the presence of an optimal balance between these two key competing actions. Next, by substituting slower moving agents with faster ones, we show that the system exhibits a performance improvement without any modifications to the original strategy. In addition, owing to the additional amount of exploitation carried out by the faster agents, we demonstrate that a heterogeneous system's performance can be further improved by reducing an agent's level of connectivity, to favor the conduct of exploratory actions. Furthermore, in studying the influence of the density of swarming agents, we show that the addition of faster agents can counterbalance a reduction in the overall number of agents while maintaining the level of tracking performance. Finally, we explore the challenges of using differentiated strategies to take advantage of the heterogeneous nature of the swarm.
\end{abstract}


%
\IEEEpeerreviewmaketitle

\section{Introduction}
With the increasing popularity of multi-robot systems (MRS), so too is the interest in implementing swarm intelligence algorithms, paving the way for swarming MRS. These systems have found multiple applications including area mapping~\cite{Kit2019}, dynamic area covering~\cite{Zoss2018}, target tracking~\cite{Esterle2020}, and performing tasks in areas that pose environmental hazards to human operators~\cite{Schranz2020}. Compared to centralized MRS, there are three main advantages in using decentralized swarming MRS: (1) flexibility---the ability for the system to swiftly adapt to rapidly changing environments, (2) robustness---the ability to cope with component failures within the system, and (3) scalability---the ability to carry out tasks in systems comprised of different number of agents~\cite{Dorigo2021}. These attributes stem from the lack of a central controller that dictates the actions of the individual agents. The decentralized nature of such systems divides up the overall control of the MRS and gives individual robots the ability to make decisions based on their local environment. This allows them to react quickly to changes in their surroundings (flexibility), and removes a central point of control, allows the system to continue operations despite agent failures (robustness) and eliminates potential processing bottlenecks (scalability).

Besides these three key features, an often overlooked attribute of swarming MRS is its \textit{modularity}. This allows for agents with different properties and behaviors to be implemented within the same collective, resulting in a heterogeneous system. Heterogeneity in swarm compositions can be utilized in several ways including employing agents with different communication ranges~\cite{Christensen2015}, or improving a system's mapping capabilities by partitioning a search-space and assigning the partitioned regions to robots according to their movement and sensing capabilities~\cite{Shi2020}. In addition, a system's modularity presents the opportunity to incrementally introduce upgrades to a fraction of agents without the need for a system-wide upgrade. Such incremental upgrades will allow for the testing of new agent functionalities at a reduced cost~\cite{Vallegra2018}. With the presence of upgraded agents, such heterogeneous swarms have the potential of delivering the same performance level with a lower agent count, thereby yielding a more cost-effective solution.

Although it has previously been shown that simply adding upgraded agents to a swarm tracking a non-evasive target can result in improved tracking performances and reduced target capture times in a target search-and-track task~\cite{Iwama2012, Kwa2020b}, this might not be the case in general. It has been shown in a wide variety of situations that changing the capabilities of agents also changes the ideal exploration-exploitation balance associated with optimal system performance~\cite{Kwa2022}. As such, when dealing with heterogeneous swarms, modifications may need to be made to the original swarm behavioral rules to account for the different exploration-exploitation balance and better take advantage of the upgraded agents. However, it is worth stressing that heterogeneity does not automatically yield an improved system's performance; not carrying out such modifications in the behavioral rules may even result in degraded system performances. For instance, in an area exploration task, it was shown that the introduction of robots with a larger sensor range resulted in higher amounts of agent clustering due to the higher amount of exploitation carried out, thereby increasing the amount of time required to explore the search space~\cite{Altshuler2009}. When determining the ideal composition of heterogeneous agents of a swarm carrying out various tasks, it was found that in some tasks, such as area coverage and foraging, the presence of different agent types resulted in poorer system performances~\cite{Nishikawa2016}.

One of the tools used to adjust the exploration-exploitation balance is by tailoring an agent's level of communication or interaction with its neighbors~\cite{Kwa2022}. In what follows, the term ``communication" has to be understood in its broadest sense, namely ``information exchange" and is in no way limited to a particular communication channel or protocol. In general, high amounts of inter-agent communication lead to higher levels of exploitation while low levels of communication yield high levels of exploration. Despite the large body of research dedicated to analyzing the effect of varying an agent's communications range on the swarm's exploration-exploitation balance and overall behavior~\cite{Coquet2019, Talamali2021}, little is known about the effect of adding one or more neighbors to an agent's communications network~\cite{Kwa2022}. Among these works, it has been demonstrated that to attain multi-agent consensus about a static noisy environment, there exists an optimum number of neighbors---and therefore an optimal exploration-exploitation balance---at which the error between the attained consensus and the ground truth is minimized~\cite{Crosscombe2021}. Similar observations have also been made in dynamic environments, where the optimum level of connectivity is related to the speed of the driving signal~\cite{Mateo2019, Kwa2021}. 

In this work, we study how heterogeneity in a swarm can be leveraged to offer adaptivity in the exploration-exploitation balance. The ultimate goal is of course to use this available adaptivity to maximize system's performance when dealing with fast-evolving circumstances. Specifically we consider a swarm attempting to search for and track multiple mobile targets. This task is carried out within the formalized framework of Cooperative Multi-robot Observation of Multiple Moving Targets (CMOMMT). Under this framework, agents position themselves in an attempt to maximize the total amount of time that each target is observed by at least one agent~\cite{Parker1997}. Most of the attention given to this problem is focused on maximizing the number of tracked targets when: (i) the latter exceeds that of the tracking agents, and (ii) when the targets travel slower than their pursuing agents~\cite{Khan2018}. Due to the assumption that fast-moving targets will always be able to evade their pursuers, cases where targets are able to travel faster and have the ability to outrun their pursuers tend to be overlooked or simply ignored. This is despite it being shown to be possible when the pursuing agents have vision of the entire environment~\cite{Janosov2017,Zhang2019}.

Given the relative lack of attention dedicated to this challenging problem of tracking multiple fast-moving evasive targets, our starting point is the strategy previously proposed in~\cite{Kwa2021}, which was designed for use in a homogeneous swarm. This approach is a combination of: (i) an adaptive inter-agent repulsive behavior, with (ii) an agent-level persistent point of attraction, and facilitated by (iii) agent-based memory. Using this strategy, the balance in the exploration of the environment and the exploitation of target information by the heterogeneous swarm can be tuned by adjusting both the memory length, as well as the degree of connectivity, $k$, of the inter-agent communications network. Tailoring the length of memory~\cite{Falcon-Cortes2019, Nauta2020c} and the number of neighbors in an agent's network~\cite{Blackwell2018, Kwa2020a} have previously been established as a method that can be used to tune this exploration-exploitation balance in various multi-agent systems. Specifically, increasing memory lengths and connectivity levels serve to increase exploitative actions while decreasing memory lengths and connectivity yields higher levels of exploration. Applying such techniques in this work allows to uncover the existence of an optimal balance between exploration and exploitation, which maximizes the system's tracking performance.

In this paper, we expand on our previous work done in \cite{Kwa2021} by focusing on the challenges and opportunities offered by a heterogeneous swarming strategy for the tracking of fast-moving evasive targets. Specifically, we investigate how the substitution of regular agents by fast-moving ones affects the swarm's exploration-exploitation balance and tracking performance. To accomplish this, we implement the aforementioned strategy and compare the performances of simulated homogeneous and heterogeneous swarms comprised of up to 50 agents while tracking up to 3 fast-moving evasive targets. The heterogeneous system is comprised of agents with two different maximum speeds. It should be emphasized that the evasive targets have a higher maximum velocity than any of the agents, including the fast-moving ones. Through our simulations of an actual robotic system, we show that altering an agent's memory length and level of connectivity confirms the presence of an optimal balance between exploratory and exploitative actions carried out by the system. Therefore, operating at this optimum maximizes the group-level tracking performance. It should be noted that the optimum engagement ratio changes depending on the task presented to the system, as well as the heterogeneous composition of the system~\cite{Kwa2020b}. In addition, we establish that smaller heterogeneous swarms are able to match the performance a full-sized homogeneous one over a range of different density of agents. Finally, we explore the challenges of using differentiated tracking strategies in heterogeneous swarms.

\section{Problem Formulation and Multi-Agent Dynamics}
\subsection{Problem Statement}
Here, we define the problem under the CMOMMT framework. We consider a set of $N$ tracking agents ${A=\{a_1, a_2, \ldots, a_N\}}$ and a set of $J$ targets ${O=\{o_1, o_2, \ldots, o_J\}}$ moving within a bounded two-dimensional square search space of dimensions $L \times L$. The agents can either be slower agents, or faster upgraded agents. Each agent is also given a memory of length, $t_{\text{mem}}$, and communicates with a fixed number of neighbors, $k \in [1, N-1]$. These aspects are discussed in the following sections. 

\subsection{Search and Track Strategy}
The search and track strategy implemented in this work was first introduced in \cite{Kwa2021} for use in homogeneous swarms tracking both evasive and non-evasive targets. This strategy consists of two components, namely `repulsion' (or avoidance) and `attraction'. The approach involves an `attraction' component that is closely tied to the topology of the interaction network, which effectively controls the flow of information throughout the system. The `repulsion' component is designed as an adaptive inter-agent repulsive behavior that yields simultaneous exploratory actions while the units are exploiting information about a tracked target. An appropriate integration of these two components promotes system exploration and exploitation respectively by generating a velocity vector by combining them at the start of each time-step:
\begin{equation}
    \bv_i[t] = \bv_{i,\text{att}}[t] + \bv_{i,\text{rep}}[t],
    \label{eqn:movement}
\end{equation}
where $\bv_{i,\text{att}}[t]$ (resp. $\bv_{i,\text{rep}}[t]$) is the velocity vector obtained from the attraction (resp. repulsion) component. In the simulation, the resultant velocity vector, $\bv_i[t]$ is then scaled by $v_{\text{max}}$, the maximum velocity of an agent, before being used to calculate the agent's position in the subsequent time-step. This strategy employed by the swarm, which is detailed in Algorithm~\ref{alg:strategy}, makes use of short-term agent memory and a variable $k$-nearest neighbor communications network to adjust the system's level of exploration and exploitation. These two parameters will be discussed in the subsequent sections. For further details regarding the individual strategy components, the reader is directed to \cite{Kwa2021}.

\begin{algorithm}
    \caption{: Dynamic $k$-Nearest Network Search and Tracking Strategy}
    \label{alg:strategy}
    \begin{algorithmic}[1]
    \STATE Set $t = 0$, $k \in [2, N-1]$, $\omega=1$, and $ c=0.5$
    \WHILE{System active}
        \FOR{All agents $i \in [1, N]$}
            \STATE Determine point of attraction
            \STATE Calculate $\bv_{\text{att},i}$ 
            \STATE Calculate $\bv_{\text{rep},i}$ 
            \STATE $\bv_i[t] \gets \bv_{\text{att},i}[t] + \bv_{\text{rep},i}[t]$ // Apply Eq~\ref{eqn:movement}
            \STATE $\bv_i[t] \gets  (v_{\text{max}}/v_i[t]) \cdot \bv_i[t]$ // Scale velocity vector with maximum agent speed
            \STATE $\bx_i[t+1] \gets \bx_i[t] + \bv_i[t]$ // Update agent position
        \ENDFOR
        \STATE $t \gets t+1$
    \ENDWHILE
    \end{algorithmic}
\end{algorithm}

\subsection{Swarm Communication Network}
The agents within the swarming system are connected to each other through the use of a dynamic $k$-nearest neighbor network. Changing the connectivity by means of the network degree, $k$, has been shown to have large impacts on the overall collective response of a swarm subjected to dynamic changes of varying time-scales~\cite{Mateo2019}. It has been demonstrated that higher levels of connectivity causes swarming agents to cluster in small areas, signifying high levels of exploitation, while lower levels of connectivity are associated with more exploratory actions~\cite{Altshuler2009, Kwa2020a}. As such, the overall exploration-exploitation balance of the swarm presented in this paper is primarily adjusted by changing the value of $k$. It is important to note that in our framework, a neighborhood is understood in the network sense, hence, an agent $i$ has as many neighbors as its degree, $k$. Note that this network is a temporal network owing to the motion of neighboring agents. Instead of modifying the collective behavior of the MRS by changing the communication range of the swarm, we designate the number of neighbors in an agent's communication network. This allows for a higher level of control over the system's exploration-exploitation balance. In practice, this can be attained by adjusting the communications range of an individual agent until the desired number of neighbors is attained~\cite{Rausch2019} or by only processing information originating from the neighboring units according to the selected value $k$ of the inter-connecting topological network~\cite{Kwa2020a}.

\subsection{Short-Term Agent Based Memory}
In \cite{Kwa2021}, we introduced a short-term agent based memory into the search and track strategy. This was inspired by the work done in \cite{Nauta2020b,Nauta2020c} that demonstrated increases in system performance when a swarm was placed in a non-destructive foraging scenario. This was especially so when resources were clustered together in the environment, allowing agents to return to resource dense patches more often. While tracking non-evasive targets, it was determined that the implementation of memory causes the swarm to aggregate in a location where the target is no longer present, thereby degrading the overall swarm's tracking performance~\cite{Coquet2019, Kwa2020a, Coquet2021}. However, in \cite{Kwa2021}, we demonstrated that the inclusion of memory was essential to tracking a fast-moving evasive target. Since an evasive target moves to avoid contact with the pursuing agents, the inclusion of memory in this scenario allowed for a more persistent point of attraction to be generated, effectively increasing the amount of exploitative actions performed by the swarm. This in turn allows the system to close in on a target even though none of the agents are able to detect the presence of a target. This can happen as long as an agent has a valid point of attraction, thereby enabling the tracking of the target by the swarm.

\section{Simulation Conditions}
\subsection{Swarm Robotic Platform}
In this work, we model the ``Bunch of Buoys" (BoB) system, originally intended for the dynamic monitoring of open water bodies~\cite{Zoss2018,Vallegra2018}. The current system employs two types of buoys: BoB-0, which weighs $7.4~\text{kg}$ and can attain a maximum speed of $1.0~\text{m/s}$, and BoB-1, weighing $3.2~\text{kg}$ and is capable of attaining a top speed of $2.6~\text{m/s}$. Being faster and lighter, BoB-1 is able to better react to dynamic changes in its operating environment compared to BoB-0. Both robotic platforms have propulsion systems allowing for omnidirectional movement. In addition, the platforms house a suite of sensors, enabling the characterization of their local environment. A distributed mesh communications system is used for sharing locally sensed information among the buoys, as well as for sending and receiving of commands. In the simulations, 50 buoys tracking a fictitious moving target were modelled. Swarms with various compositions of BoB-0 and BoB-1 units and levels of connectivity were tested to observe the effects of gradually adding upgraded units to the system.

\subsection{Simulation Parameters and Performance Metric}
The bounded two-dimensional square search space of dimensions $L \times L$ is such that $4 \leq L \leq 445$. The slow and fast agents have associated maximum velocities of $\bv_{a_0,\text{max}} = 0.1$, $\bv_{a_1,\text{max}} = 0.26$ respectively and $x$ and $y$ positions represented by ${\bx_i = (x_i, y_i)}$. The agent's memory of length is given as $t_{\text{mem}} \in [0, 50]$. In simulations involving differentiated communications strategies, $k_f$ (resp. $k_s$) represents the number of topological neighbors for fast (resp. slow) agents.

The targets are modeled using disc-shaped binary objective functions with fixed radii of ${\rho = 1}$. These targets also have $x[t]$ and $y[t]$ positions associated with the center of the disc and a maximum velocity, $\bv_{o, \text{max}}$, and do not overlap. This maximum velocity is greater than that of the two types of agents, therefore: ${\bv_{a_0, \text{max}} < \bv_{a_1, \text{max}} < \bv_{o, \text{max}}}$. A target is considered to be tracked if there exists an agent within its radius. Formally:
\begin{equation}
\label{eqn:coverage}
    \text{cov}(o_m, t)=
    \begin{cases}
        1 & \exists i \in A \text{ s.t. } \|\bx_i - \bx_m\| \leq \rho, \\
        0 & \text{otherwise.}
    \end{cases}
\end{equation}
The modeling of the target as a binary objective function is done to make the task even more challenging and is similar to a visual search task where a target needs to be seen to have its presence confirmed~\cite{Esterle2020}. This is in contrast to tracking a target from the intensity of an emitted signal (e.g. radio signal strength, chemical plume, etc.) in which various gradient-descent methods can be used. Indeed, such gradient-descent methods form one of the most widely used techniques to track targets and become completely ineffective when dealing with such binary objective functions. When using a binary objective function as done here, it ensures that such simplistic techniques are not used by the system. It can be argued that the full power of swarm intelligence can only be accessed when dealing with such challenging scenarios.

The targets are set to move according to an evasive movement policy. Specifically, a targets initially travels towards randomly generated waypoints within the search space until it encounters a swarming agent within its radius. Upon this encounter, the target calculates its velocity using the repulsion Eq.~\eqref{eqn:rep}: 
\begin{equation}
    \bv_{\text{rep},m}[t] = - \sum_{i\in \mathcal{N}_i}\left( \frac{a_{R}}{r_{mi}[t]}\right)^d \frac{{\mathbf{r}_{mi}[t]}}{r_{mi}[t]},
    \label{eqn:rep}
\end{equation}
where $\mathbf{r}_{mj}[t]$ is the vector from target $m$ to agent $i$ at time-step $t$, with agent $i$ being a part of $\mathcal{N}_i$, the group of agents located within the target's radius. The magnitude of the repulsion vector is determined by $r_{mi}[t]$, where $r_{mi}[t] = \lVert \mathbf{r}_{mi}[t] \rVert$, and $a_{R}$, the repulsion strength of the target.

Should the target encounter agents within its radius for $t_{\text{limit}}$ consecutive iterations, the target travels in a straight line for $t_{\text{evade}}$ time-steps in an attempt to outrun its pursuers.

The goal of the system is to maximize its tracking performance of the targets within this environment, which can be measured quantitatively by means of the normalized reward function:
\begin{equation}
\label{eqn:cost_fn}
    \Xi = \frac{1}{T J} \sum^T_{t=1}\sum^J_{m=1}\text{cov}(o_m, t),
\end{equation}
where $T$ it the total time period of interest and $J$ is the total number of targets within the search space. 

\subsection{Exploration and Exploitation Balance}
\label{sec:eed_bal}
In addition to group-level tracking performance, the overall complex collective dynamics of a swarm can be studied through the quantification of the exploration-exploitation balance of the system. This metric was first introduced in \cite{Kwa2021}. Here, the proportion of a swarm's agents engaged with a target, which shall be referred to as the swarm's \textit{engagement ratio}, is used to quantify the balance between exploratory and exploitative actions carried out by the system. An agent is considered to be engaged with a target if it has entered the `tracking' state and $S_i(\bx_i[t], t) = 1$, thereby giving the overall normalized engagement ratio:
\begin{equation}
    \label{eqn:engagement}
    \Theta = \frac{1}{NT}\sum^T_{t=1}\sum^N_{i=1} S_i[t].
\end{equation}
When using this metric, systems that display a high engagement ratio indicates that agents are spending a high fraction of time attempting to track a target, i.e., exploiting it. Conversely, a low engagement ratio indicates that the agents are spending less time attempting to track a target, which is characteristic of higher levels of exploration. It should be noted that while our initial work in \cite{Kwa2021} pointed to the existence of an optimum engagement ratio to maximize tracking performance, this optimum changes according to the task assigned. As such, it would not be possible to declare what an optimum level of engagement should be. This changing optimum engagement ratio will be addressed in the subsequent sections.

\section{Simulation Results}
Swarms of $N=50$ agents comprised of various proportions of fast and slow agents are deployed within the square search space. The width of the search space is kept constant at $L=30$ for all simulations except for those studying swarm density, where $4 \leq L \leq 445$. The agents are tasked with tracking $J \in \{ 1, 2, 3 \}$ targets following the evasive movement policy that traveled at a maximum speed of $\bv_{o, \text{max}} = 0.3$ non-dimensional units per time-step. All simulations are carried out over a period of 400,000 iterations to ensure the statistical stationarity of the results. 

\subsection{Impact from the Level of Network Connectivity}
\label{sec:connectivity}
\begin{figure}[htbp]
\centering
\includegraphics[width=\textwidth]{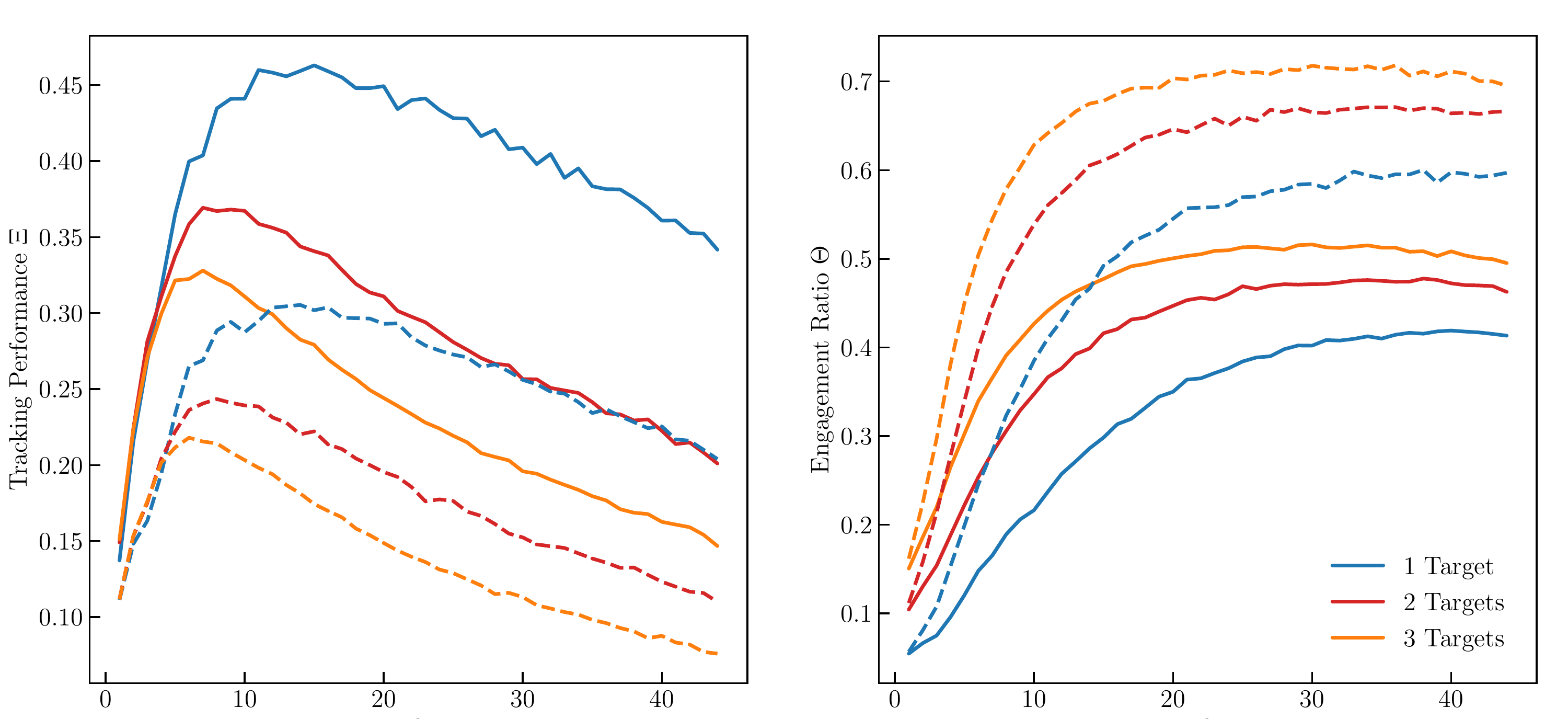}
\caption{Tracking performance (right) and engagement ratio (left) of a heterogeneous swarm comprised of 45 fast agents and 5 slow agents (solid line), as well as a homogeneous swarm comprised of 50 slow agents (dashed line) tracking multiple fast-moving evasive targets. Simulations were performed using various levels of network connectivity $k$ and a memory length of $t_{\text{mem}}=20$.}
\label{fig:multi_target_k}
\end{figure}

\begin{figure}[htbp]
\centering
\includegraphics[width=0.6\textwidth]{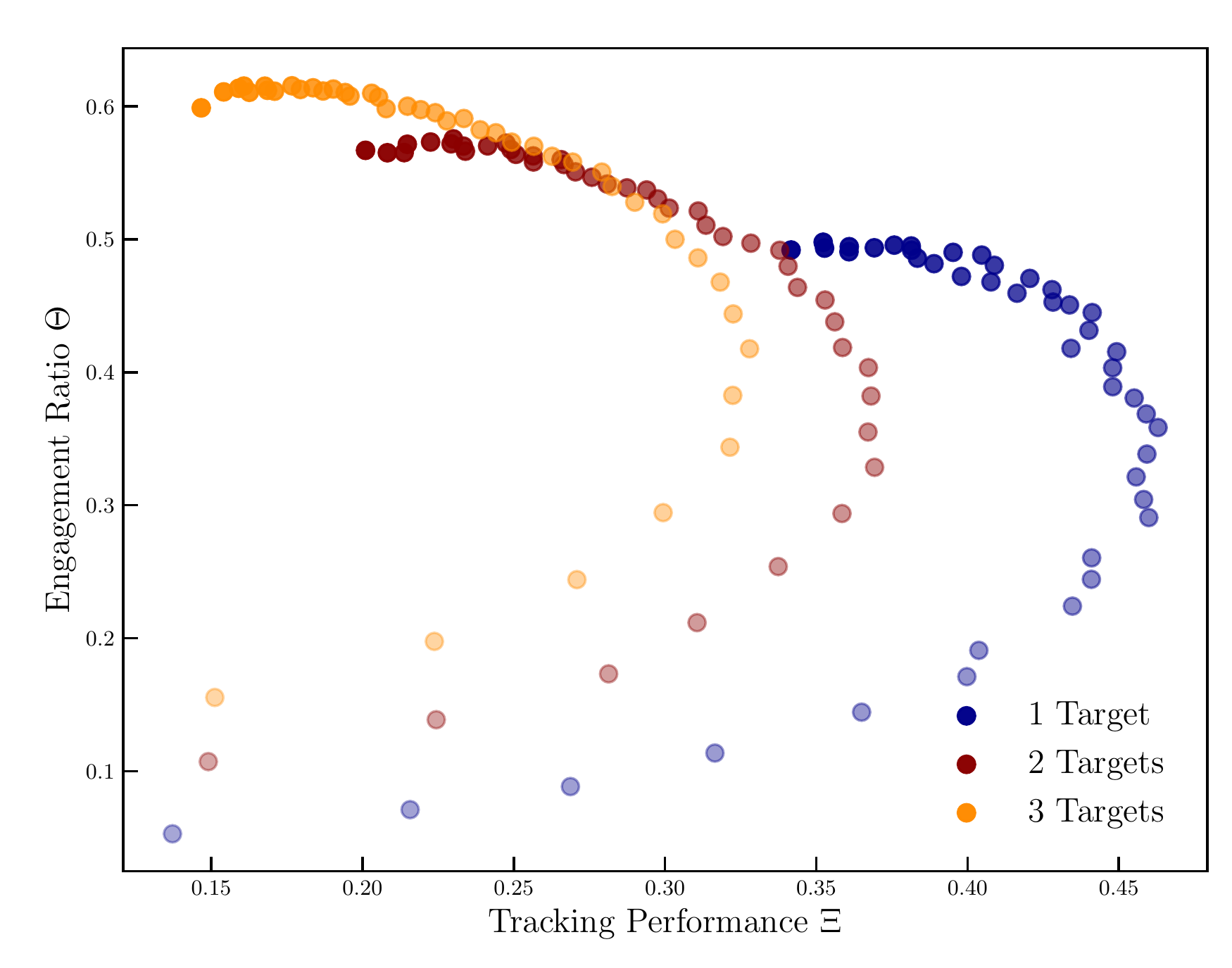}
\caption{Engagement-Tracking plots of a heterogeneous swarm comprised of 15 fast agents tracking multiple fast-moving evasive targets. Darker shaded points indicate swarms communicating through networks with higher values of $k$.}
\label{fig:multi_target_explore_exploit}
\end{figure}



Figure~\ref{fig:multi_target_k} illustrates that there exists an optimum level of connectivity, $k^*$, at which the tracking performance of the system is maximized. This is true for both homogeneous and heterogeneous swarms. On one hand, when $k<k^*$, the low tracking by the system is due to the low level of exploitation, as attested by the low values of $\Theta$, resulting in the agents being unable to cluster around the target adequately. On the other hand, when $k>k^*$, too much exploitation carried out by the system yields excessive clustering of agents around the target, ultimately impairing the tracking performance. Indeed, when operating at high levels of $k$, the agents take a longer time to re-explore the environment and reacquire a target's position after being outrun. The figure also suggests that this optimum $k$ is reduced when fast agents are introduced to the swarm. This implies that while no modifications to the overall approach are necessary to improve the swarm's performance when introducing fast agents, doing so by altering an agent's connectivity degree will maximize the performance of the heterogeneous swarm. The reasons for this will also be further explored in Sec.~\ref{sec:fast_agents}. 

Secondly, Fig.~\ref{fig:multi_target_k} reveals that while tracking evasive targets, the strategy considered in homogeneous swarms can readily be used with a heterogeneous one to a greater effect without the need for adjustments; an improvement in the MRS's tracking performance occurs regardless of the system's level of connectivity. This result is congruent with our previous findings that no specific modifications needed to be made to benefit from a heterogeneous swarm in a dynamic area coverage scenario~\cite{Vallegra2018}. This is despite the fact that other studies have shown that either modifications to the overall strategy, such as changes to the communications network, need to be made, or that specific initialization conditions need to be satisfied to fully benefit from a heterogeneous system~\cite{Altshuler2009, Nishikawa2016}. The reasons for the improvement in tracking performance in spite of the change in exploration-exploitation balance are explained in Sec.~\ref{sec:fast_agents}.

Thirdly, it is demonstrated that the tracking performance of the swarm decreases when more targets are added to the environment. Again, this occurs regardless of the system's level of connectivity. This is to be expected as simultaneously tracking multiple targets is more challenging than just tracking a single one. It can also be observed that both increasing the system's degree of connectivity and increasing the number of targets also serves to raise the engagement ratio, $\Theta$. The former occurs because elevating the system's level of connectivity results in a target's positional information being spread around the system's agents more readily. The latter is due to the higher number of targets naturally increasing the likelihood of a tracking agent encountering a target. Consequently, in these two situations, the likelihood of an agent attempting to engage with a target also increases, either due to the increased availability of target information or higher probability of an agent encountering a target, thereby increasing $\Theta$. It should be noted that $\Theta$ does not differentiate between targets. 

Interestingly, the optimum exploration-exploitation balance appears to remain constant regardless of the number of targets---rightmost point in Fig.~\ref{fig:multi_target_explore_exploit}. Therefore, to maximize the tracking performance of the swarm, more exploration needs to be carried out. As more exploration is carried out at lower levels of connectivity, a swarm's $k^*$ reduces when it attempts to track multiple targets, which can be seen in Fig.~\ref{fig:multi_target_k}.

\subsection{Impact from Agent-Based Memory}
\begin{figure}[htbp]
\centering
\includegraphics[width=0.6\textwidth]{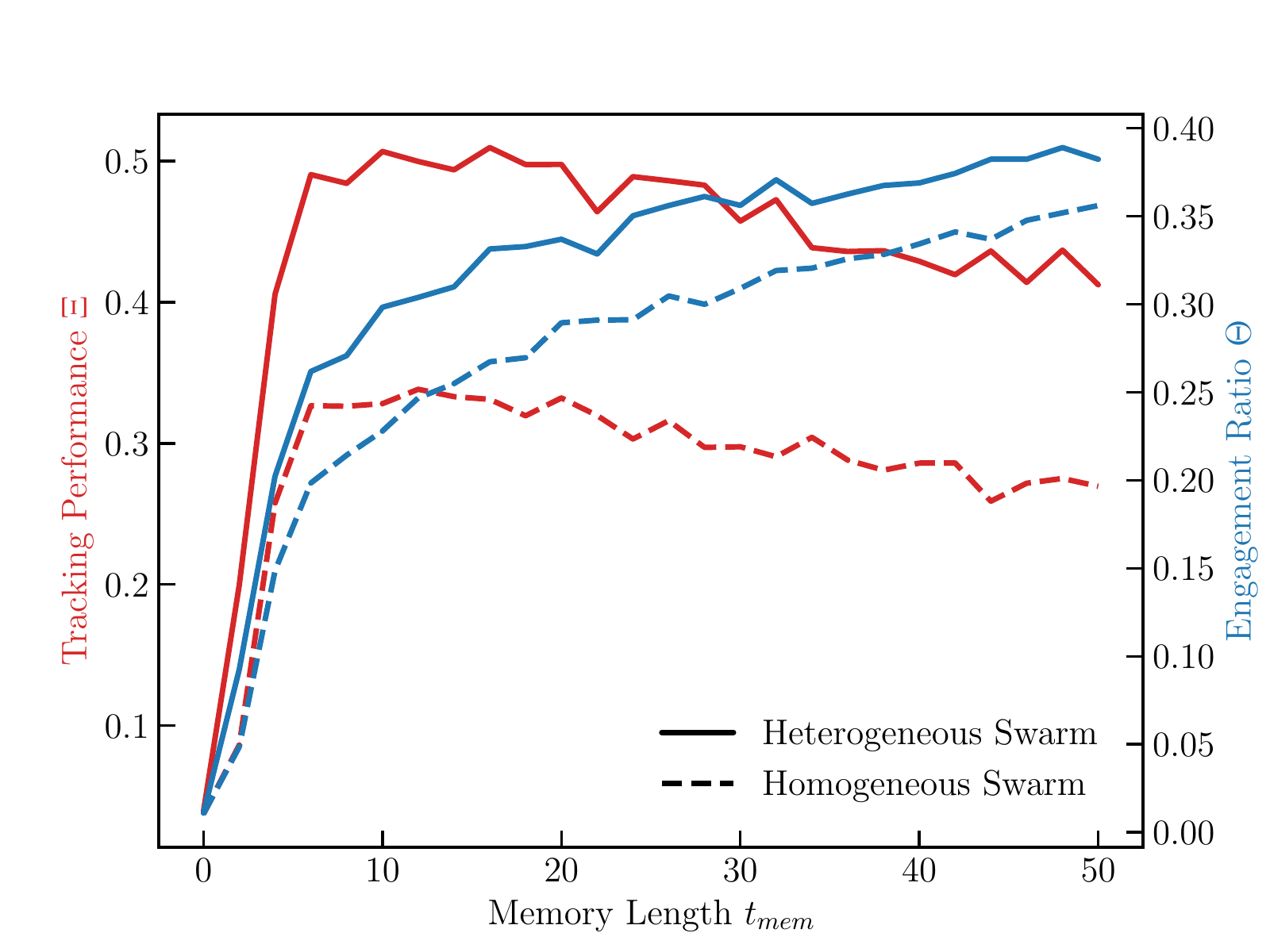}

\caption{Tracking performance (red line) and engagement ratio (blue line) of a heterogeneous swarm comprised of 15 BoB-1 units and 35 BoB-0 units (solid line) and a homogeneous swarm comprised of slow agents (dashed line) tracking a single fast-moving evasive target. Simulations were performed using a network connectivity of $k=14$ and various memory lengths.}
\label{fig:memory}
\end{figure}


As previously mentioned, the implementation of an agent-based memory was done to create a more lasting point of attraction, thus allowing the agents to flock around an evasive target. Since it facilitates the clustering of agents around the target, it also functions as another tool that permits the adjustment of the exploration-exploitation balance of the swarm. However, caution needs to be taken when implementing agent memory as its use was previously determined to be detrimental when tracking non-evasive targets due to the system's use of information that is no longer valid when excessive memory lengths were used. 

Despite its associated disadvantages, it can clearly be seen in Fig.~\ref{fig:memory} that the use of a short-term memory is essential to the swarm's ability to track an evasive target. Without the use of memory ($t_\text{mem} = 0$), the system is unable to effectively track an evasive target and the addition of small amounts of memory is associated with a sharp increase in both tracking performance and engagement. This is because allowing the agents to exploit information from its memory generates a persistent point of attraction based on a target's last known position. Without the use of memory, even though the swarm may periodically encounter a target, the swarm tends to expand until the agents reach their static equilibrium positions and is unable to close in on the target because of its quick evasive maneuvers. 

However, should the agent be allowed to exploit information from a longer-than-optimal memory length, there is an overall degradation of the tracking performance as the system starts to suffer from the effects stemming from the exploitation of information that is no longer valid. This is despite the fact that the engagement ratio of the system continues to rise with the increasing memory lengths. This rise in $\Theta$ is due to the increased amount of time the swarming agents spend in the `tracking state' (see Sec.~\ref{sec:eed_bal}), causing them to aggregate at a position where a target is no longer present. This further emphasizes the importance of maintaining an optimal balance between exploration and exploitation as mentioned in Sec.~\ref{sec:connectivity}.

\subsection{Impact from the Introduction of Fast Agents}
\label{sec:fast_agents}

\begin{figure}[h]
\centering
\includegraphics[width=\textwidth]{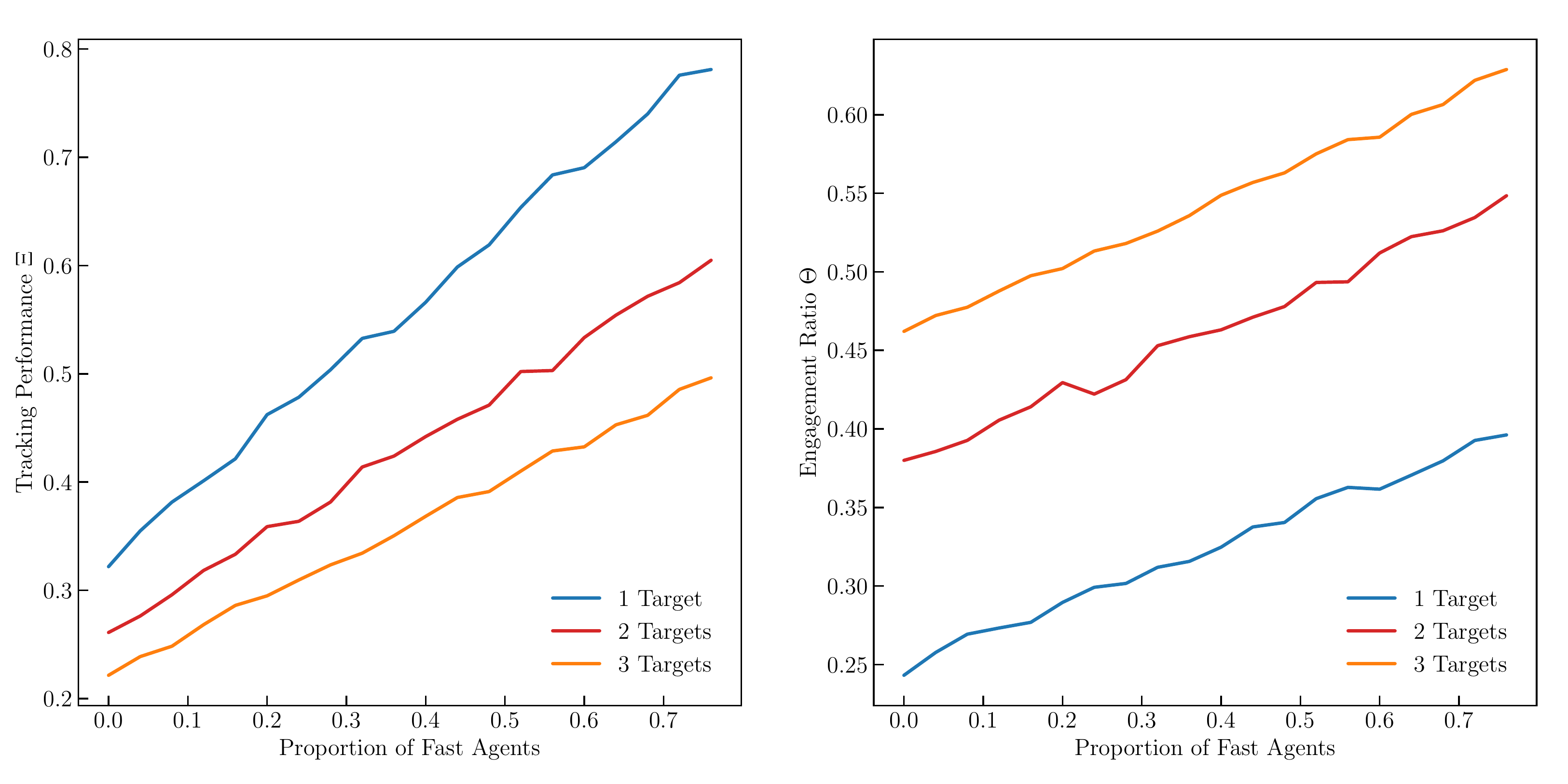}
\caption{Tracking performance (right) and engagement ratio (left) of swarms comprised of 50 agents with various proportions of BoB-1 agents tracking a fast-moving evasive target. Simulations were performed using a network connectivity of $k=12$ and a memory length of $t_{\text{mem}}=20$.}
\label{fig:multi_target_proportion}
\end{figure}

\begin{figure}
\centering
\includegraphics[width=\textwidth]{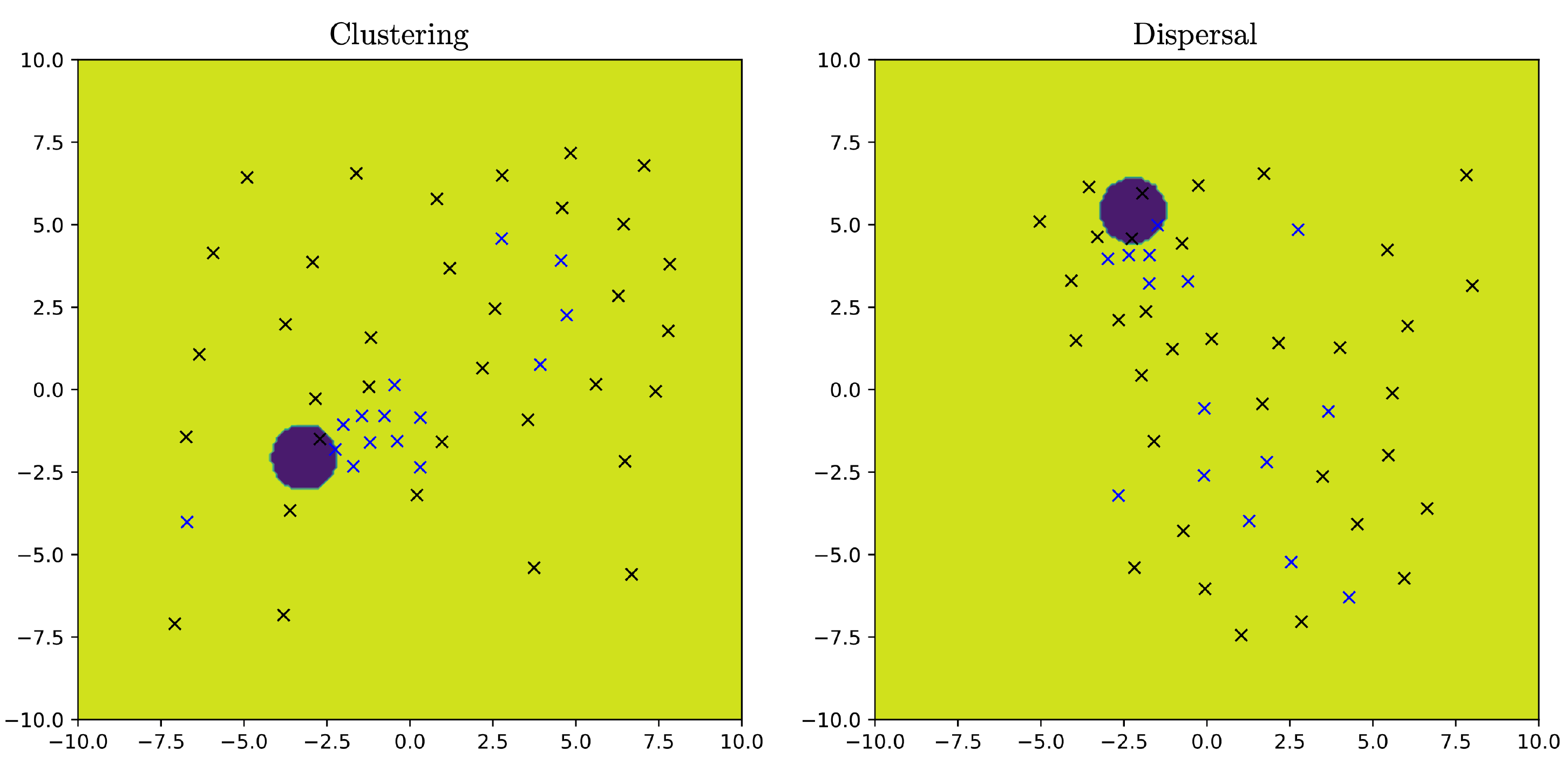}
\caption{Heterogeneous swarm comprised of 35 BoB-0 units (black crosses) and 15 BoB-1 units (blue crosses) tracking an evasive target (dark blue disc). (Left) BoB-1 units clustering while pursuing target; (Right) BoB-1 units redistributed around the environment due to inter-agent repulsion.}
\label{fig:clustering}
\end{figure}

As can be seen in Fig.~\ref{fig:multi_target_proportion}, increasing the proportion of fast agents results in an increase in both $\Xi$ and $\Theta$. It is worth highlighting again that these increases are obtained without modifying the original tracking strategy. Observations made from the simulations reveal the reasons why such modifications are unnecessary. As illustrated in Fig.~\ref{fig:clustering}, while the faster agents may flock together when actively trying to engage with the target, they are eventually able to disperse and distribute themselves across the swarm after losing track of the target. This is thanks to the adaptive inter-agent repulsion component in the search and track strategy, which increases the chances of a fast agent encountering a target.

\begin{figure}
\centering
\includegraphics[width=\textwidth]{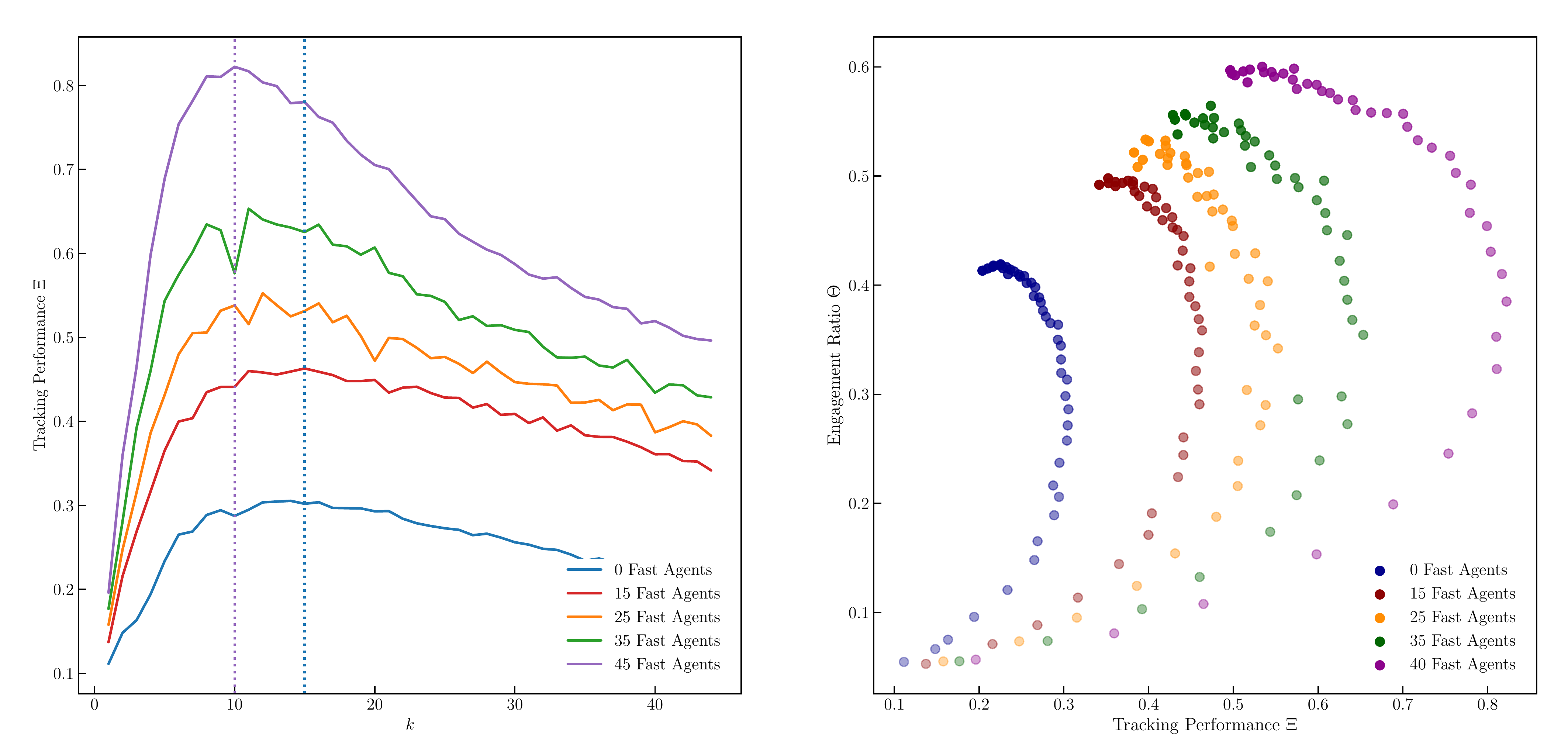}
\caption{Swarms comprised of 50 agents with different agent compositions of BoB-1 and BoB-0 units tracking a single fast-moving evasive target. A system with 0 fast agents is a homogeneous swarm comprised solely of BoB-0 units. (Left) Tracking performance of the system while operating with different levels of connectivity $k$. (Right) Engagement-Tracking plots where darker shaded points indicate swarms communicating through networks with higher values of $k$.}
\label{fig:explore_exploit}
\end{figure}

The observations and the quantitative results suggest that the increases in performance and engagement are solely due to the increased mobility of the faster agents within the swarm. Since the faster agents are better able to keep up with the targets, they directly increase the swarm's tracking capability. They are also able to disseminate more information about the targets' locations across the swarm, allowing neighboring agents to remain engaged with a target for longer and close in on it, further improving the tracking performance. The increase in $\Theta$ also means that the addition of faster agents biases the system's actions towards higher levels of exploitation. However, due to the increased mobility of the system in general, the optimum exploration-exploitation balance of the swarm also moves in favor of exploitation. This is illustrated in Fig.~\ref{fig:explore_exploit} where gradually increasing the number of fast agents within the system results in an increased optimum engagement ratio.

Despite this apparent shift in the ideal exploration-exploration balance in favor of exploitation, Fig.~\ref{fig:explore_exploit} also shows that the optimum connectivity degree, $k^*$, appears decreases with the introduction of fast-moving BoB-1 units into the system---from $k^* \sim 15$ when a homogeneous swarm is used to $k^* \sim 10$ when the swarm consists of 45 BoB-1 units and 5 BoB-0 units. This reduction in $k^*$ is characteristic of a swarm trying to increase its level of exploration. This observation is similar to the results obtained when tracking non-evasive targets~\cite{Kwa2020b}, albeit to a lesser extent. This suggests that the primary effect of substituting in fast agents into a swarm is the increase of both the system's capability and propensity to exploit target information. To balance the increased amount of exploitation promoted by the faster agents, the system needs to carry out more exploration. As more exploration is carried out at lower levels of connectivity, the optimum connectivity degree, $k^*$, decreases when faster agents are substituted into the system.

\subsection{Impact of Swarm Density}
\begin{figure}
\centering
\includegraphics[width=\textwidth]{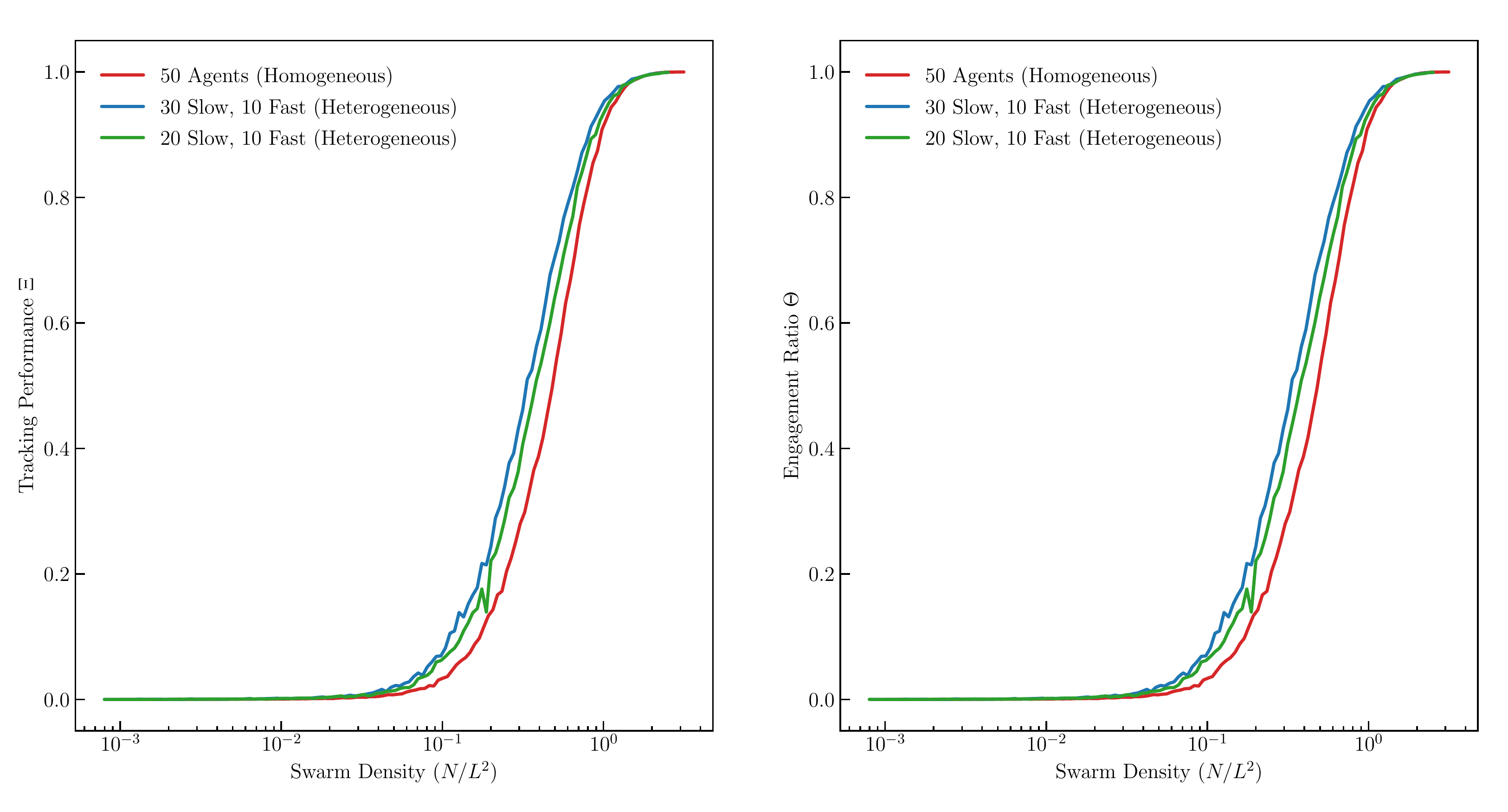}
\caption{Evasive tracking performance (left) and engagement ratio (right) of a homogeneous swarm comprised of 50 agents and two heterogeneous swarms comprised of a total of 40 and 30 agents each. Simulations were carried out over various area sizes to vary swarm density with agents using a $k=14$ network.}
\label{fig:het_density}
\end{figure}

As alluded to previously, the inclusion of highly motile agents within the swarm can also reduce the overall number of agents required in the system. However, such a reduction in agent numbers will result in the swarm operating with a reduced swarm density. As shown in Fig.~\ref{fig:het_density}, when varying the density of agents, three different phases can be observed. At low density ($<3\times10^{-2}$ agents / unit area), the swarm is unable to effectively track the target. In this phase, the swarm density is too low and agents in the system are unable to collectively follow the target. At high density ($>1$ agent / unit area), the system is able to constantly track the target as it moves around the search environment. In this phase, the search area is too small and the target will encounter an agent regardless of where it moves. Between these two phases is a `transition' region where the tracking performance sharply increases with density. 
The figure also reveals that within this transition region, heterogeneous swarms that include faster moving agents are able to match, if not exceed, the tracking ability of a homogeneous swarm solely comprised of non-upgraded agents in all three density phases. This is because the faster agent speeds are able to compensate for the reduced agent numbers. Since the faster agents are able to better keep up with the target, the swarm is able to remain engaged with the target, allowing for an adequate amount of exploitation to be performed. As such, a larger swarm comprised of slower agents is able to outperform a smaller swarm that includes faster agents. 

\subsection{Heterogeneous Strategies}

\begin{figure}[htbp]
\centering
\includegraphics[width=0.6\textwidth]{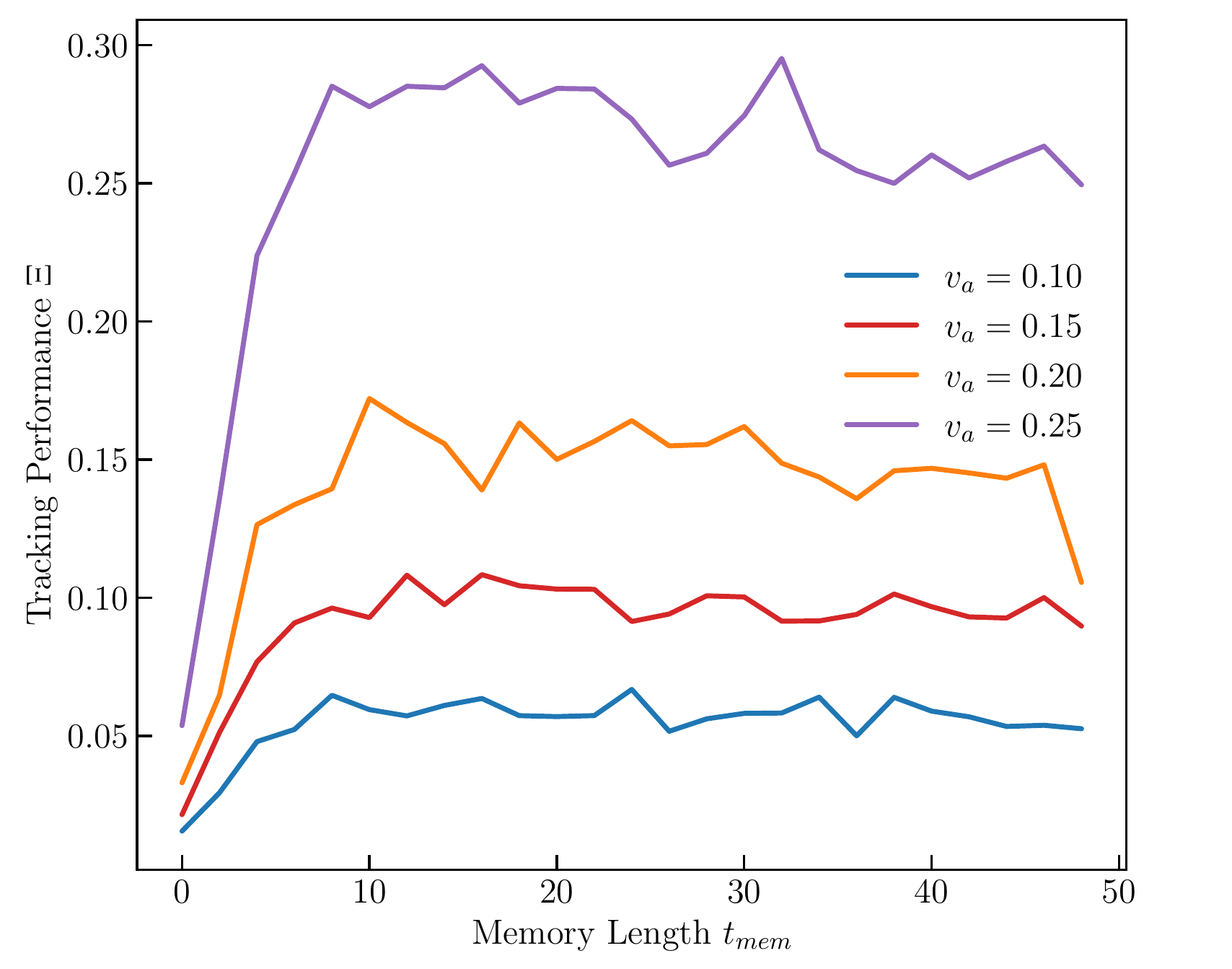}
\caption{Evasive tracking performance of homogeneous swarms with different $v_a,\text{max}$ and memory lengths. Simulations were run at $k=14$ and $v_{o,\text{max}} = 0.3$}
\label{fig:memory_speed}
\end{figure}

To ascertain the possibility of using a heterogeneous memory strategy, simulations are carried out using homogeneous swarms with varying agent speeds while utilizing different memory lengths. However, as shown in Fig.~\ref{fig:memory_speed}, the optimum memory length does not appear to change with the maximum speed of the agents. This therefore suggests that the use of a heterogeneous strategy involving the use of different memory lengths will not improve the tracking performance of a swarm. 

We demonstrated in Sec.~\ref{sec:fast_agents} that the addition of faster moving agents reduces the optimal level of connectivity, $k^*$, for the swarm. In addition, previous studies have shown that this optimal level of connectivity changes with the speed of the driving stimulus~\cite{Mateo2019, Kwa2020a}. As such, to explore the possibility of using a heterogeneous strategy using different levels of connection, we carried out the same evasive target tracking simulation where the level of connectivity of the fast agents ($k_f$) is different from the level of connectivity of the slow agents ($k_s$). 

\begin{figure}[htbp]
\centering
\includegraphics[width=0.6\textwidth]{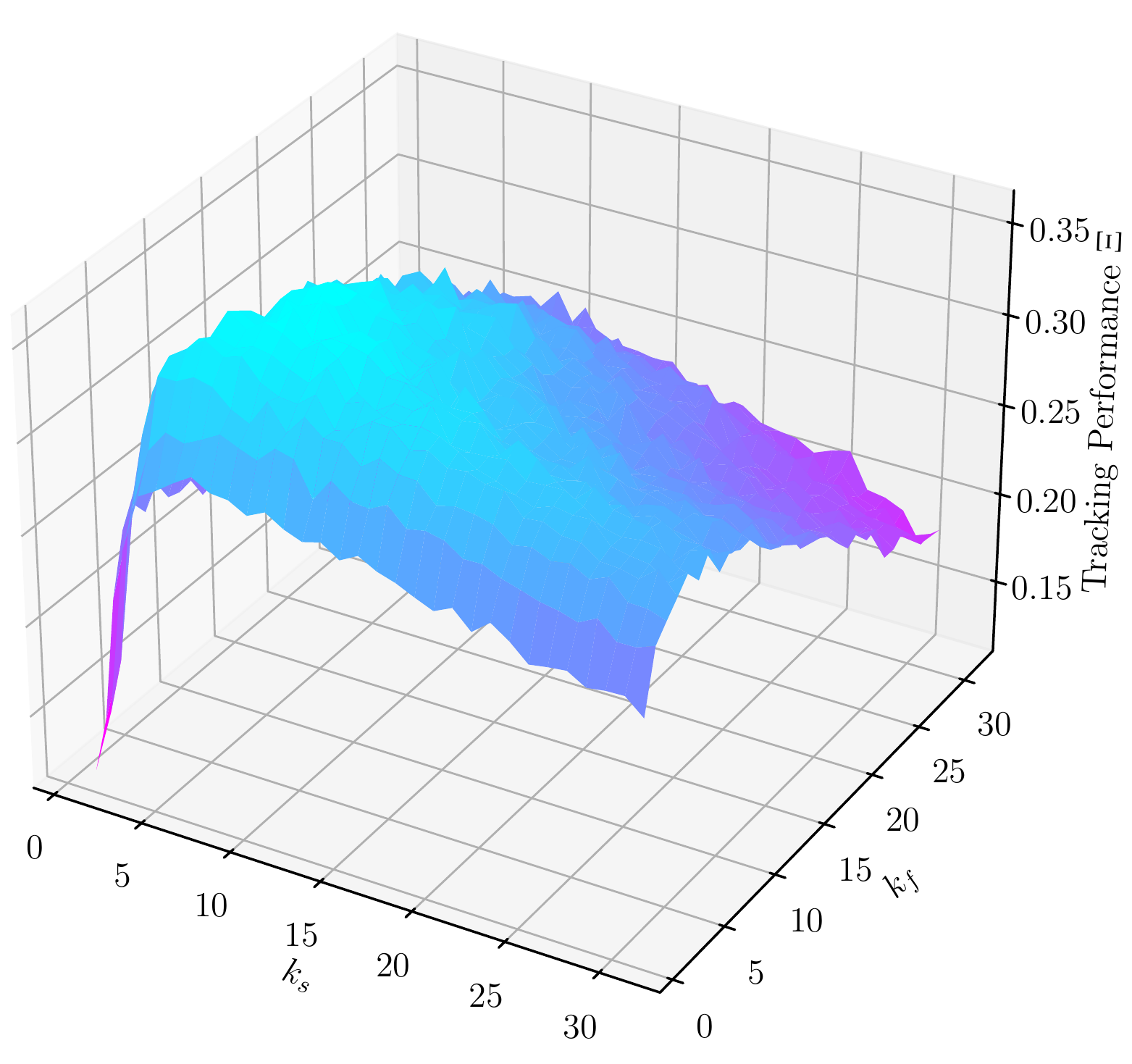}
\caption{Single evasive target tracking performance of a heterogeneous comprised of 30 slow agents and 10 fast agents with slow and fast agents, each having different levels of connectivity.}
\label{fig:variable_k}
\end{figure}

From Fig.~\ref{fig:variable_k}, we were able to identify an optimum combination of levels of connectivity to maximize the tracking performance. However, this optimum occurs when $k_s = k_f = 10$ and none of the other cases tested suggest that employing a differentiated strategy would further improve a swarm's tracking performance. These results hint that further attempts to improve the tracking performance of heterogeneous MRS swarms should integrate other behavioral factors: e.g., differentiated task allocation based on agent's placement, prediction of target movement patterns, etc. 

\section{Discussion}
The use of decentralized swarming MRS offers users many advantages, including system modularity. By exploiting this modular character, heterogeneous swarms comprised of agents with different behaviors and capabilities can be implemented, potentially leading to improved system performance. However, the different agent abilities may need to be taken into account when designing swarming strategies due changes in the ideal exploration-exploitation balance.

In this work, we studied the implementation of heterogeneous swarms performing the task of searching for and tracking multiple fast-moving evasive targets under the CMOMMT framework. Through simulating an actual environmental monitoring robotic system employing swarming agents, we demonstrated that the introduction of faster agents is indeed able to improve the overall tracking performance of the swarm without modifying the originally implemented behaviors. This is because the higher mobility of the faster units gives them the ability to better keep up with and track the targets, leading to better information dissemination of the targets' locations and higher levels of engagement with the target. This also leads to an overall shift in the optimum exploration-exploitation balance of the system, favoring more exploitative behaviors to maximize the swarm's tracking performance. While the strategy does not need to be modified, the performance of the swarm can be further improved by accounting for the increased amount of exploitation brought on by the introduction of fast-moving agents. By reducing the agents' degree of connectivity, more exploratory action is promoted within the system, thereby offsetting the excessive exploitation brought on by the introduction of the fast agents. This allows for the ideal exploration-exploitation balance to be restored, thereby improving the performance of the swarm. It should be noted that while we have shown that the optimum level of connectivity, $k^*$, decreases with the proportion of fast agents within the swarm, a more detailed study would be required to explore how $k^*$ varies as a function of the swarm composition.

Through varying the size of the operating environment, we have also showed that smaller heterogeneous swarms incorporating faster agents are able to match and outperform the performance of a full-sized homogeneous swarm comprised at all swarm densities. This is because faster agent speeds are able to track the evasive target for longer periods of time. This compensates for the reduced agent numbers and allows the swarm to remain engaged with the target, allowing for an adequate amount of exploitation to be performed. 

We have also demonstrated the impact of the level of network connectivity and the introduction of agent-based memory on the swarm's tracking performance, as well as their effects on the level of exploration and exploitation carried out by the swarm. By increasing either the level of connectivity, $k$, or the length of agent memory, the system biases itself towards exploitation. Conversely, reducing $k$ or memory length causes the swarm to favor more exploratory actions. As such, there exists an optimal level of connectivity and memory length at which there is a suitable balance between exploration and exploitation carried out by the system. This balance is responsible for maximizing the tracking performance of the swarm. This optimum moves in favor of exploration when tracking more than one target to counter the higher levels of target engagement provoked by the presence of multiple targets. Beyond this analysis of the influence of $k$, the degree of the network topology, it has recently been shown that behavioral changes in a swarm can propagate according to simple or complex contagions depending on other network parameters such as network clustering or Kirchhoff index~\cite{horsevad2022transition}. Looking into the influence of these other network characteristics might reveal other parameters that influence the swarm's response when tracking targets.

It should be noted that in this work, we have not touched on an agent's communication range, which may pose an issue when operating at very low swarm density in expansive environments. In such scenarios, the lack of communication neighbors would hinder the amount of exploitation carried out by the system, undermining its ability to track a target. Similarly, it was also assumed that agents would have perfect information of the target's location when within the target radius and that there was no noise affecting an agents sensing capabilities. As can be expected, the addition of noise would also affect the agents' ability to exploit a target's positional information. Future works on this subject should therefore focus on trying to encourage and control the level of exploitation when the range of communication is restricted or when imperfect data is provided to the agents.

Given the ability to tune the balance between exploratory and exploitative actions in a heterogeneous swarm, further explorations of this topic may offer new ways to learn collective behaviors through multi-agent reinforcement learning (MARL) that could potentially take advantage of differentiated strategies for fast and slow agents~\cite{Kouzehgar2020}. While this work performed a preliminary study on the use of differentiated strategies in heterogeneous swarms where agents had different levels of mobility, a system's heterogeneity can come in different forms (e.g., deploying agents with different sensing abilities, mixed aerial and ground robot swarms, etc.). Further work should be done on the application of differentiated strategies for such systems, i.e. behavioral heterogeneity.





\ifCLASSOPTIONcaptionsoff
  \newpage
\fi

\end{document}